\documentclass[sn-aps]{sn-jnl}
\catcode`\|=12\relax 
\usepackage{orcidlink} 
\usepackage[T1]{fontenc}
\usepackage{graphicx}
\usepackage{tikz} 
\usepackage[]{framed}
\usepackage{amsmath}
\usepackage{adjustbox}
\usepackage{xspace}
\usepackage{soul}
\usepackage{placeins}  
\usepackage{tcolorbox}
\tcbuselibrary{breakable}
\usepackage{float}
\usepackage[inline]{enumitem}
\usepackage{booktabs}
\usepackage{array, multirow, multicol}
\usepackage{geometry}
\newcommand\thedataset{\textsc{PublicHearingBR}\xspace}
\newcommand\cemmilpodcasts{\textsc{Cem Mil Podcasts}\xspace}
\newcommand\xlsum{\textsc{XL-Sum}\xspace}
\newcommand\wikilingua{\textsc{WikiLingua}\xspace}
\newcommand\recognasumm{\textsc{RecognaSumm}\xspace}
\newcommand\temario{\textsc{TeMário}\xspace}
\newcommand\cstnews{\textsc{CSTNews}\xspace}
\newcommand\rulingbr{\textsc{RulingBR}\xspace}

\tcbset{
    style_prompt/.style={
        fontupper=\small
    }
}

\begin{document}

\title[\thedataset: A Brazilian Portuguese Dataset of Public Hearing Transcripts for Summarization of Long Documents]{\thedataset: A Brazilian Portuguese Dataset of Public Hearing Transcripts for Summarization of Long Documents}

\author*[1]{\fnm{Leandro} \sur{Carísio Fernandes},\orcidlink{0000-0002-4114-2334}}\email{carisio@gmail.com}
\author[2]{\fnm{Guilherme} \sur{Zeferino Rodrigues Dobins},\orcidlink{0009-0002-4336-2143}}
\author[2]{\fnm{Roberto} \sur{Lotufo},\orcidlink{0000-0002-5652-0852}}
\author[3]{\fnm{Jayr} \sur{Alencar Pereira},\orcidlink{0000-0001-5478-438X}}

\affil*[1]{\orgname{Câmara dos Deputados}, \orgaddress{\city{Brasília}, \country{Brazil}}}
\affil[2]{\orgname{Universidade Estadual de Campinas (Unicamp)}, \orgaddress{\city{Campinas}, \country{Brazil}}}
\affil[3]{\orgname{Universidade Federal do Cariri (UFCA)}, \orgaddress{\city{Juazeiro do Norte}, \country{Brazil}}}

\abstract{This paper introduces \thedataset, a Brazilian Portuguese dataset designed for summarizing long documents. The dataset consists of transcripts of public hearings held by the Brazilian Chamber of Deputies, paired with news articles and structured summaries containing the individuals participating in the hearing and their statements or opinions. The dataset supports the development and evaluation of long document summarization systems in Portuguese. Our contributions include the dataset, a hybrid summarization system to establish a baseline for future studies, and a discussion of evaluation metrics for summarization involving large language models, addressing the challenge of hallucination in the generated summaries. As a result of this discussion, the dataset also includes annotated data to evaluate natural language inference tasks in Portuguese.}


\maketitle

\section{Introduction}

With huge volumes of text being added daily to information systems, it is difficult and time-consuming for users to search through this material, which often includes a mix of relevant and irrelevant information \cite{ats_comprehensive_review, ats_scientific_papers_transformers}. Thus, in many contexts, it is more practical to access a summary that contains the most relevant information from the data.

Text summarization is a natural language processing task that reduces a text to a shorter version while retaining its main ideas. This task is important for various applications, including information retrieval, content management, and automatic news article writing, among others. Since manual summarization is time-consuming and often impractical, developing automatic systems to perform this task is necessary \cite{ats_comprehensive_review}.

Automatic text summarization techniques are generally divided into extractive, abstractive, and hybrid methods \cite{automatic_summarization,survey_text_summarization}. Abstractive summarization generates new sentences that capture the essence of the original text, while extractive summarization selects and concatenates segments directly from the source. Hybrid methods combine elements of both strategies. Traditionally, most research has focused on extractive methods. However, in recent years, the emphasis has shifted to abstractive or hybrid approaches \cite{ats_comprehensive_review}. With advancements in generative artificial intelligence systems, these methods have become increasingly relevant.

Documents involved in summarization tasks vary widely, ranging from short news articles to lengthy research papers and even collections of multiple documents. Each type presents distinct challenges. Summarizing short documents, such as news articles, is easier to perform and evaluate. Multi-document summarization typically depends on a prior retrieval stage focused on a specific topic within those documents. Summarizing long documents usually involves high compression rates, requiring careful assessment of the most relevant pieces of information. As the information may be dispersed throughout the text, this task can be complex.

A typical pipeline for evaluating an automatic text summarization system includes several stages: pre-processing, summarization, post-processing, and evaluation \cite{survey_text_summarization}. The system is evaluated by comparing the generated summary with a reference summary using metrics such as ROUGE \cite{lin-2004-rouge} and Check-Eval \cite{checkeval}. Regardless of the metric, publicly available standardized datasets are generally used to evaluate new methods.

The availability of datasets is essential to advancing research and development in summarization methods \cite{Koh_2022}, as they provide the necessary resources to train and evaluate new systems. However, most available datasets are in English, which limits the development of summarization tools for non-English speakers.
Portuguese, one of the world's most widely spoken languages\footnote{\url{https://en.wikipedia.org/wiki/List_of_languages_by_total_number_of_speakers}}, illustrates this situation. Despite its wide use, there is a lack of summarization datasets in Portuguese, and existing ones usually focus on short documents like news articles, leaving a gap in resources for long document summarization.

To address this gap, this paper introduces \thedataset, an open dataset for summarizing long documents in Brazilian Portuguese. The documents are transcripts of public hearings held by the Brazilian Chamber of Deputies, during which individuals discuss a specific topic. Each hearing's transcript is published online, and certain characteristics of these documents may present challenges for summarization, including:
\begin{enumerate*}
	\item They are extensive texts, often spanning dozens of pages and sometimes exceeding one hundred;
	\item The opinions of a given individual may be dispersed throughout the text;
	\item Some opinions of a participant may be inferred from their agreement or disagreement with others; and
        \item The topics discussed are diverse, ranging from socially controversial issues (e.g., vaccination mandates, minority rights, and climate change) to highly specific topics (e.g., regulation of online platforms and installation of offshore wind farms).
\end{enumerate*}

\thedataset consists of 206 pairs of transcripts with corresponding news articles and metadata that indicate the participants in the hearing and their opinions (a structured summary). The articles were extracted from the \textit{Agência Câmara de Notícias}\footnote{\url{https://www.camara.leg.br/noticias}}, the news agency of the Brazilian Chamber of Deputies. Thus, we consider these articles to be adequate summaries of the hearings. The structured summaries were extracted from the articles.

To test the dataset, we used ChatGPT with a custom prompt\footnote{\url{https://openai.com/index/introducing-gpts/}} to extract relevant information from the transcripts using a hybrid summarization approach. The goal is not to solve the problem of summarizing public hearings, but to establish a baseline using an off-the-shelf tool available to end users. To evaluate the results, we propose using large language models (LLMs) to compare the experiment’s output with the dataset. We use LLMs both to verify whether the extracted opinions are present in the dataset (similar to the Check-Eval metric \cite{checkeval}) and to identify hallucinations.

The discussion on hallucinations also led to manually annotated data containing 4\,238 opinions, excerpts of text, and a flag indicating whether the opinions can be inferred from those excerpts. In other words, \thedataset can also be used as a natural language inference (NLI) dataset for Brazilian Portuguese.

The contributions of this paper are:
\begin{enumerate*}
    \item a dataset with 206 pairs of long documents and summaries in Brazilian Portuguese, designed for long document summarization;
    \item an LLM-based hybrid summarization system, which provides a baseline for future studies;
    \item a discussion of evaluation metrics for summarization using LLMs, including how to identify hallucinations in the generated summaries;
    \item additional annotated data derived from the previous items, consisting of 4\,238 opinions, text excerpts, and a manually annotated flag indicating if the opinions can be inferred from those excerpts, enabling its use for NLI tasks in Brazilian Portuguese.
\end{enumerate*}

\section{Related Work}

The availability of datasets in Portuguese is limited, especially for text summarization. Although large corpora in Brazilian Portuguese have been published recently (e.g., \cite{ulyssses_tesemo}), they are intended for other purposes, such as fine-tuning LLMs. Existing summarization datasets in Portuguese are scarce and usually focus on short texts.

One of the first summarization datasets available in Portuguese is the \temario \cite{temario}, which contains 100 pairs of journalistic news articles from two newspapers, along with summaries written by a specialist. On average, the texts contain approximately 600 words, and the summaries about 200 words. The specialist was explicitly instructed to keep the summaries between 25\% and 30\% of the original text length. The dataset was later expanded with an additional 150 samples, with an average of 1\,200 words per article \cite{temario_2006}.

Other Portuguese datasets for news summarization include the \cstnews \cite{cst_news} and the \recognasumm \cite{recogna_summ}. \cstnews is a manually annotated dataset with 140 short texts. All the texts together amount to about 47\,000 words -- that is, just over 330 words per sample. On the other hand, \recognasumm contains about 130\,000 samples and considers the summary to be the heading and subheading of the news. However, these elements often describe the main topic of the news to capture the reader's attention, rather than serving as an effective summary of the article.

News articles are, in general, short texts. One Portuguese dataset with slightly longer documents is the \rulingbr \cite{ruling_br}, which contains over 10\,000 rulings from the Brazilian Supreme Federal Court. Each document includes four sections: \textit{Ementa}, \textit{Relatório}, \textit{Voto}, and \textit{Acórdão}. The \textit{Ementa} section is considered the reference summary of the decision. Documents average around 1\,200 words and can reach up to 2\,000 words. Summaries contain up to 150 words, averaging about 90 words, which corresponds to approximately 7\% of the total document length. This dataset focuses on summarizing judicial texts to extract the court’s decision.

There are also multilingual summarization datasets, notably \wikilingua \cite{wiki_lingua} and \xlsum \cite{xl_sum}. \wikilingua focuses on ``How To'' texts, while \xlsum targets news articles. Both provide texts in English paired with summaries written in one or more languages (up to 18 languages for \wikilingua and up to 44 for \xlsum). In general, the texts are short. For instance, \wikilingua articles average 391 tokens, and their summaries 39 tokens, based on the mBART tokenizer \cite{liu2020multilingual}.

Recently, the \cemmilpodcasts dataset was released for non-commercial research. It contains transcripts and some metadata for about 114\,000 podcasts in Brazilian Portuguese and 8\,000 episodes in European Portuguese. On average, transcripts have about 9\,500 words, with a median of approximately 6\,700 words. The longest transcript has about 205\,000 words. The task is to generate a description for each episode using the transcript as input and compare it to the one provided by the podcast creator. Access to the dataset requires permission from the maintainers. However, according to the official website\footnote{\url{https://podcastsdataset.byspotify.com/}}, as of December 2023, they are no longer accepting access requests due to shifting priorities.

Despite the availability of some summarization datasets in Portuguese, there is a lack of open datasets for summarizing long documents. This is a gap that \thedataset fills. It includes 206 long documents and their summaries. The documents are transcripts of public hearings held by the Brazilian Chamber of Deputies on various topics, and the task is to extract the relevant individuals and their opinions. The summaries consist of metadata indicating the individuals and what they said or supported, along with a related news article about the public hearing. The dataset is open and can be freely accessed\footnote{\url{https://huggingface.co/datasets/unicamp-dl/PublicHearingBR}}. Besides, \thedataset differs from other Portuguese datasets not only because of the length of the documents to be summarized, but also because the summaries list the main points raised during the hearings rather than being limited to a sentence or a single decision.

\section{\thedataset -- A long document summarization dataset}

\subsection{Concept and purpose of the dataset} \label{sec:concept_and_purpose}

The committees of the Brazilian Chamber of Deputies conduct public hearings on various topics of national interest. The transcripts of these hearings are lengthy public documents, often spanning dozens or even a hundred pages, and are freely available on the official website\footnote{\url{https://www.camara.leg.br}}.

\textit{Agência Câmara de Notícias}, the news agency of the Brazilian Chamber of Deputies, regularly publishes news articles about these hearings. The articles, which are short documents, can be seen as summaries of longer ones -- the transcripts. From them, we can extract structured summaries (metadata) indicating the main topic of the hearing, the participants in the debate, and their opinions or statements.

\thedataset groups this information into triples, each containing a public hearing transcript, a news article, and metadata linking the opinions to the individuals who participated in the hearing. As the article was written by a specialized team, we consider both its text and the metadata extracted from it as ground truth (gold standard). 

Given the characteristics listed above, \thedataset can be used for the following tasks:

\begin{enumerate}

    \item Summarizing long documents to extract the most relevant opinions of the individuals: the transcript can be used to identify the main participants in the public hearing and their opinions, which can then be compared with the metadata provided in the dataset.
    
    \item Writing a news article from the metadata: the metadata can serve as input for a pipeline that writes an article, which can be compared to the one provided in the dataset.
    
    \item Writing a journalistic article using a long document: the transcript can be used to write a news article, which can be compared with the one available in the dataset.
    
\end{enumerate}

The experiments presented in this paper focus on the first use case of the dataset: summarizing long documents to extract the most relevant opinions of the individuals involved. The other two use cases involve news generation and are outside the scope of this article.

\subsection{Step-by-step process for creating \thedataset} \label{sec:step_by_step}

\begin{figure}[htbp]
    \centering
    \includegraphics[width=1\linewidth]{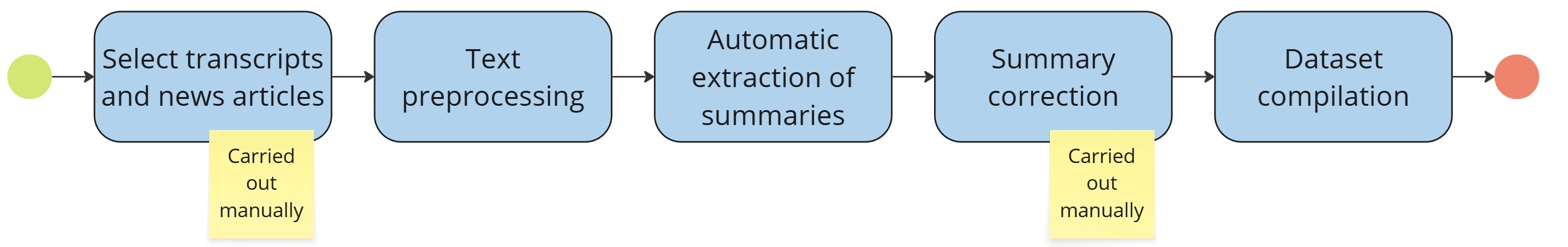}
    \caption{Flowchart for the creation of \thedataset.}
    \label{fig:fig_dataset_step_by_step}
\end{figure}

Figure \ref{fig:fig_dataset_step_by_step} shows the flowchart for the dataset creation process. It began with the manual selection of URLs for news articles published by \textit{Agência Câmara de Notícias} related to public hearings. In this step, we selected 206 articles covering hearings held between November 2021 and May 2024. Then, we manually searched the websites of the committees that conducted the hearings to find the URLs of their transcripts.

The next step was to download the HTML files for both the selected articles and the corresponding transcripts and parse them. During post-processing, the HTML structure was removed to extract just the main text.

The raw text of the articles was used to extract the structured summary. This process could have been done entirely manually. However, to speed it up, we opted to use an LLM to extract a preliminary summary, which was then manually corrected. We used OpenAI's GPT-4o model with the prompt shown in Figure \ref{fig:prompt_extract_metadata_from_articles}\footnote{All prompts used in this article are available in Appendices \ref{sec:prompts_pt} (original version, in Portuguese) and \ref{sec:prompts_en} (translated version, in English).}. The generated metadata includes the main topic of the hearing, the participants, and their opinions. At the end of this stage, considering all 206 news articles, a total of 3\,054 opinions expressed by the hearing participants were extracted.

To ensure the quality of the dataset, we manually reviewed all extracted opinions by reading the news articles and comparing them with the generated summary. Two main types of errors were found in the automatic extraction. The first was the attribution of statements to the wrong participant. The second was the splitting of a single opinion into two, which artificially inflated the total number of opinions. This happens because, in this type of news article, it is common to present a claim followed by a direct quote that illustrates it. All these errors were manually corrected. In cases of incorrect attributions, the opinions were reassigned to the correct participant. When a single opinion was split into two, they were merged.

At the end of this process, 2\,203 opinions related to the 206 dataset samples remained. This data was made available in a single JSON file\footnote{\texttt{PublicHearing\_LDS.jsonl}, available in the repository of this article.}.

\subsection{Characteristics of \thedataset}\label{sec:characteristics_dataset}

\thedataset contains 206 samples. Each sample includes a public hearing transcript (long document), a news article (unstructured summary), and a dictionary with metadata that links opinions to individuals as presented in the article (structured summary). Table \ref{tab:statistics_of_the_dataset} shows statistics on word counts in the transcripts and articles, as well as the number of individuals and opinions in the metadata.

\begin{table}[htbp]
\caption{Statistics of \thedataset.}
\centering
{%
\begin{tabular}{@{}lccccc@{}}
\toprule
	Statistic
	& \begin{tabular}[c]{@{}c@{}}Transcript \\(words)\end{tabular} 
	& \begin{tabular}[c]{@{}c@{}}Article \\(words)\end{tabular}
 	& \begin{tabular}[c]{@{}c@{}}Relative size of article\\ compared to the \\transcript (\%)\end{tabular}
	& \begin{tabular}[c]{@{}c@{}}Individuals \\(count) \end{tabular}
	& \begin{tabular}[c]{@{}c@{}}Opinions \\(count)\end{tabular} \\ 
\midrule

Mean        & 18\,102   & 627       & 4.20  & 5.2   & 10.7 \\
Std         & 12\,137   & 160       & 2.06  & 1.8   & 4.5 \\
Min         & 4\,437    & 288       & 0.75  & 2     & 3 \\
Q (5\%)     & 8\,495    & 399       & 1.76  & 3     & 5 \\
Q (25\%)    & 12\,079   & 522       & 2.75  & 4     & 8 \\
Median      & 16\,424   & 607       & 3.97  & 5     & 10 \\
Q (75\%)    & 20\,858   & 706       & 5.24  & 6     & 13 \\
Q (95\%)    & 30\,761   & 932       & 7.21  & 8.75  & 18.75 \\
Max         & 147\,728  & 1\,215    & 15.07 & 12    & 31 \\ \bottomrule
\end{tabular}%
}
\vspace{0.05cm}
\label{tab:statistics_of_the_dataset}
\end{table}

On average, the transcripts have about 18\,000 words, with 90\% of them ranging between approximately 8\,500 and 30\,500 words. Two transcripts have over 50\,000 words, and the longest has nearly 150\,000 words.

The articles have a more uniform length, averaging 627 words, with the longest one being roughly twice as long. The low variability in length is typical of news articles of the same type.

An average article is typically about 4\% of the length of an average transcript, resulting in a compression rate of about 96\%. However, due to the relatively consistent length of articles and the varying length of transcripts, an article can be up to 15\% of the length of a transcript.

On average, each article in the dataset presents the opinions of about five individuals, with roughly ten opinions per article. The maximum number of opinions in an article is 31.

\section{Experiment -- Summarization of public hearing transcripts using ChatGPT}

\subsection{Method: Description and evaluation of the experiment}
\label{sec:experiment_method}

This experiment addresses the primary use case of the dataset, as described in Section \ref{sec:concept_and_purpose}: summarizing long documents to extract the most relevant opinions of the individuals.

As shown in Section \ref{sec:characteristics_dataset}, the transcripts in the dataset are long documents and thus can exceed the context window of some LLMs. Instead of building a custom pipeline to handle this, we tested a general-purpose, off-the-shelf generative AI tool: ChatGPT, given its widespread use. This approach simulates how a typical user might handle the task with general-purpose software and also establishes a baseline for the dataset, which can be improved as more specialized systems become available.

We used ChatGPT with a instruction prompt (Figure \ref{fig:prompt_instruct_custom_gpt}) that guided the model on how to read and interpret the transcripts. The prompt specified the input data format and instructed the language model to extract a structured summary of the most relevant opinions from the transcript.

Once ChatGPT has extracted the structured summaries for all samples, they need to be compared with the ground truth provided in the \thedataset dataset to assess the quality of the summarization.

Since this summarization approach is hybrid, using the ROUGE metric \cite{lin-2004-rouge} is inadequate as it relies on n-grams rather than semantic content. We propose that the evaluation covers three aspects:
\begin{enumerate*}
    \item recall: to measure the percentage of relevant opinions the experiment was able to retrieve; 
    
    \item precision: to determine the percentage of retrieved opinions that are relevant; 
    
    \item hallucination: to ensure that the extracted opinions are present in the transcript, even if they are not in the dataset.
\end{enumerate*}

\subsubsection{Recall and precision}

We use the following steps to calculate recall and precision:

\begin{enumerate}
    \item First, individuals mentioned in the transcript are mapped to those in the metadata available in \thedataset. Since names might differ (due to abbreviations or titles, for example), a language model is initially used to perform this mapping. This was done using GPT-4o with the prompt shown in Figure \ref{fig:prompt_match_individuals}.

    \item Next, for each individual identified in the previous step, the opinions extracted from the transcript are mapped to those in the \thedataset dataset. Because these are natural language texts, it is important to determine whether opinions are semantically equivalent. This mapping was also performed using GPT-4o with the prompt shown in Figure \ref{fig:prompt_match_opinions}.

    \item Once the mapping is complete, the number of matching opinions between the extracted data and the dataset can be counted, allowing the calculation of recall and precision.

\end{enumerate}

Recall is a relevant metric because it measures the proportion of reference opinions -- identified by human annotators in the dataset -- that are successfully retrieved by the summarization system. It is worth noting that the recall calculated through the mapping performed by the LLM is the essence of the Check-Eval method. The main difference is that Check-Eval automatically generates the reference checklist from a text using an LLM \cite{checkeval}, while this method obtains the checklist directly from the dataset.

Precision is another important metric to consider, although low precision is not necessarily a negative outcome. For example, a journalist might omit some relevant opinions due to space constraints. However, it is important to assess if the system is hallucinating. This is especially important because the goal is to write a news article (or to generate inputs for one), where false information is unacceptable. Besides, since participants in public hearings share their ideas and opinions, the opinions must be attributed to the correct individuals.

Thus, low precision associated with hallucinations undermines the credibility of the system. On the other hand, low precision without hallucinations is less problematic: as long as the system sufficiently summarizes the transcript, it still offers the user a broad range of opinions to write a news article.

\subsubsection{Hallucination}
\label{sec:hallucination_calculation}

To detect hallucinations in the summarization of a transcript, we follow the steps below:

\begin{enumerate}
    \item For each transcript, separate the speeches of each person;

    \item Split each person's speech into small chunks. In this article, the text was split into chunks of five sentences with a one-sentence overlap using the spaCy\footnote{\url{https://spacy.io/}} framework and the \texttt{pt\_core\_news\_sm} model;

    \item Create a vector database for each person using these chunks. We used OpenAI's \texttt{text-embedding-3-small} model to encode the chunks;
    
    \item For each opinion in the structured summary produced by the experiment, search the corresponding individual's vector database to find the most relevant chunks. Opinions were also encoded using OpenAI's \texttt{text-embedding-3-small} model;

    \item Send the opinion extracted by the summarization and the four most relevant chunks to an LLM to check if the generated information can be inferred from the chunks. Three prompts, detailed in Section \ref{sec:results_discussion}, were tested in this step.
    
\end{enumerate}

Using this strategy, it is possible to summarize a transcript by extracting the opinions of the individuals participating in the hearing and to check if the summarization system captures most of the relevant opinions (recall). It also allows us to check if the system generates many opinions that are not relevant (precision). Finally, it is possible to identify and remove most of the hallucinations.

\subsection{Results and discussion}
\label{sec:results_discussion}

\subsubsection{Precision and recall}

In this experiment, we prompted ChatGPT to generate a structured summary (metadata) that includes individuals and their opinions. Considering all the transcripts in the dataset, a total of 4\,238 opinions were extracted.

Table \ref{tab:results} shows the recall and precision statistics from our experiment on the 206 dataset samples. The average recall was 44.9\%, meaning the ChatGPT retrieved nearly half of the relevant opinions for each sample. On the other hand, the average precision was 24.8\%, which indicates that, for every relevant opinion generated, it also produced roughly three other opinions.

\begin{table}[htbp]
\caption{Statistics of recall and precision of the results of the experiment.}
\centering
{%
\begin{tabular}{@{}lccc@{}}
\toprule
	Statistic
	& \begin{tabular}[c]{@{}c@{}}Recall (\%)\end{tabular} 
	& \begin{tabular}[c]{@{}c@{}}Precision (\%)\end{tabular} \\ 
\midrule
Mean        & 44.9  & 24.8 \\
Std         & 22.5  & 13.1 \\
Min         & 0.0   & 0.0  \\
Q (5\%)     & 11.5  & 5.6  \\
Q (25\%)    & 28.6  & 15.2 \\
Median      & 44.4  & 24.0 \\
Q (75\%)    & 60.0  & 33.1 \\
Q (95\%)    & 87.1  & 47.5 \\
Max         & 100.0 & 71.4 \\ \bottomrule
\end{tabular}%
}
\vspace{0.05cm}
\label{tab:results}
\end{table}

\subsubsection{Hallucination}

\paragraph{Manual annotation}

To assess the level of hallucination in the generated opinions,  we first manually annotated the generated content to identify hallucinations, in order to validate the automatic annotation method. To do so, we applied the first four steps described in Section \ref{sec:hallucination_calculation}  to retrieve the four most similar text chunks for each of the 4\,238 opinions generated in the experiment. A consultant from the Chamber of Deputies then reviewed each opinion and its four corresponding chunks, indicating whether the opinion could be inferred from them. If so, the opinion-chunks pair was annotated as a valid opinion; otherwise, as a possible hallucination.

The manual annotation showed that 3\,734 opinions are valid (i.e., can be inferred from the four closest chunks), while 504 (11.89\%) are possible hallucinations. This percentage represents an upper bound and may be an overestimate of the actual number of hallucinations generated by the experiment, as some opinions might be supported by text outside the four closest chunks. From now on, ``possible hallucinations'' will be referred to simply as ``hallucinations''.

We understand that this manual annotation of 4\,238 samples indicating whether an opinion can be inferred from text excerpts can also be used to evaluate NLI systems in Portuguese. Therefore, we also provide this data as a JSONL file\footnote{\texttt{PublicHearingBR\_NLI.jsonl}, available in the repository of this article.}.

\paragraph{Automatic annotation}

After the manual annotation, we applied the final step of the method to detect hallucinations: we used an LLM to check if an opinion could be inferred from the four closest chunks of text. We tested three prompts with four LLMs: GPT-4o mini, GPT-4o, DeepSeek-V3, and Sabiá-3.1 (a Brazilian Portuguese model). All prompts request the same information but phrase the request differently:

\begin{itemize}
    \item Prompt 1 (Figure \ref{fig:prompt_1_hallucination}) asks if the opinion can be fully inferred from the chunks and requests the answer in JSON format, including the reasoning behind the decision.
    
    \item Prompt 2 (Figure \ref{fig:prompt_2_hallucination}) builds on Prompt 1 and explicitly requests the sentences that support the decision before providing the reasoning. It also states that the model is an expert in text analysis.

    \item Prompt 3 (Figure \ref{fig:prompt_3_hallucination}) extends Prompt 2 by reinforcing the request and specifying that, in case of doubt, the model should conclude that the opinion cannot be inferred from the chunks. It also defines the model as an expert in discourse analysis.
\end{itemize}

\paragraph{Results and discussion of the automatic hallucination evaluation}

For each of the three prompts and four LLMs tested for automatic hallucination evaluation, we generated a confusion matrix comparing the output with the 4\,238 manually annotated opinions (3\,734 valid opinions and 504 hallucinations). Tables \ref{tab:results_gpt_4o_mini} to \ref{tab:results_sabia} in Appendix \ref{sec:confusion_matrices} show the matrices.

\begin{figure}[htbp]
    \centering
    \includegraphics[width=1\linewidth]{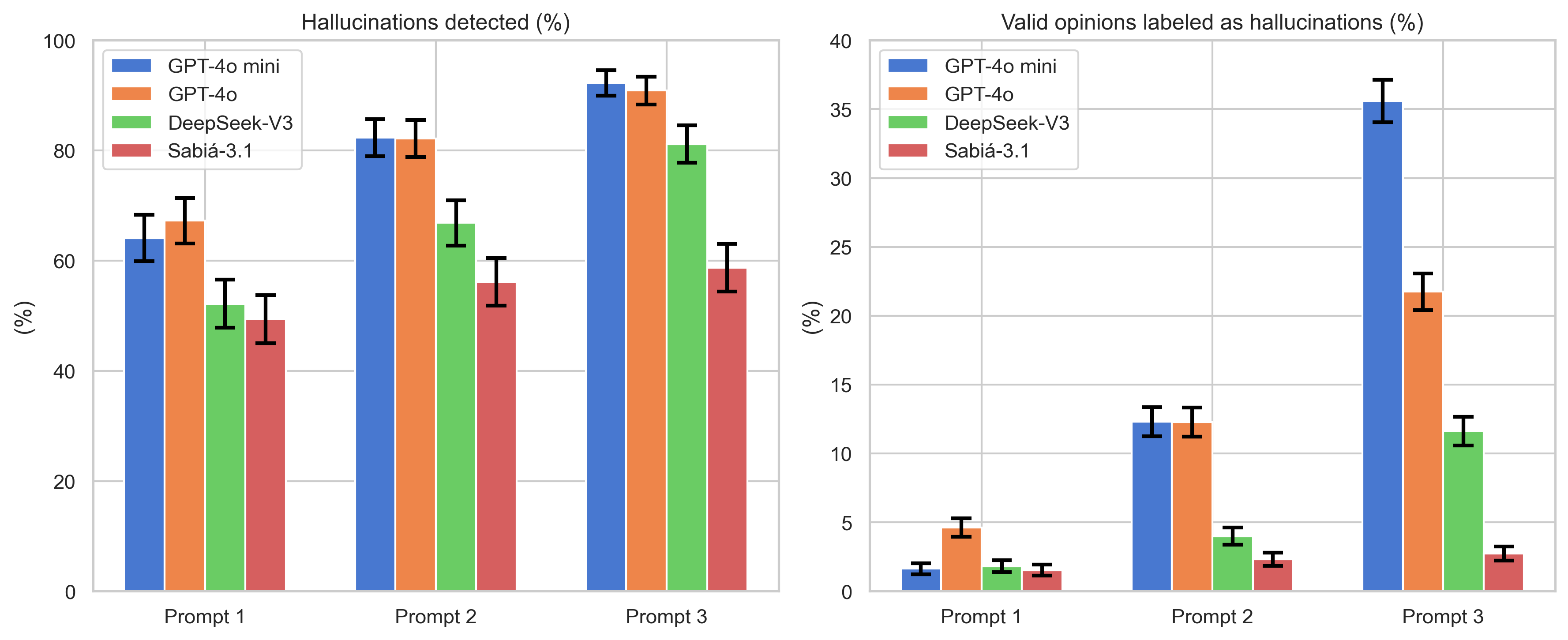}
    \caption{(Left) percentage of correctly detected hallucinations and (right) false hallucinations for the three prompts tested with GPT-4o mini, GPT-4o, DeepSeek-V3, and Sabiá-3.1 models.}
    \label{fig:fig_results_hallucinations}
\end{figure}

\begin{figure}[htbp]
    \centering
    \includegraphics[width=1\linewidth]{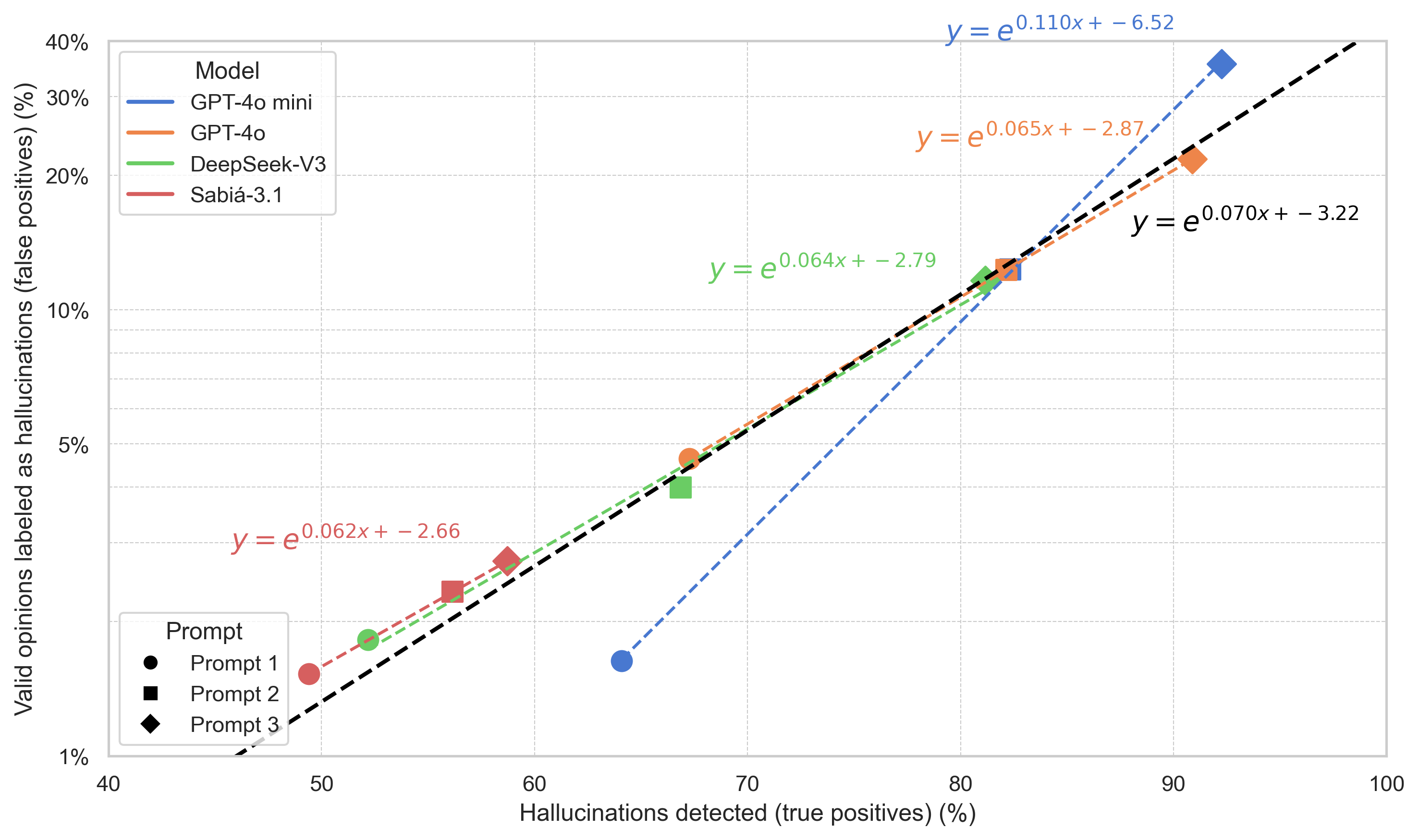}
    \caption{Trade-off between hallucinations detected (true positives) and valid opinions tagged as hallucinations (false positives) across models and prompts. The dashed lines show the linear regression for each model (colors) and across all points (black).}
    \label{fig:fig_results_true_positive_vs_false_positve}
\end{figure}

Tables \ref{tab:consolidated_hallucination_results_true_positive} and \ref{tab:consolidated_hallucination_results_false_positive} provide an overview of each model and prompt's performance in terms of correctly detected hallucinations (true positives) and valid opinions labeled as hallucinations (false positives). Figure \ref{fig:fig_results_hallucinations} shows the same information, including a 95\% confidence interval. Finally, Figure \ref{fig:fig_results_true_positive_vs_false_positve} illustrates the trade-off between the percentage of true positives and false positives across models and prompts.

\begin{table}[htbp]
    \centering
    \caption{Hallucinations correctly identified by models and prompts (true positives). Percentages are based on the 504 manually annotated hallucinations.}
    \label{tab:consolidated_hallucination_results_true_positive}
    \begin{tabular}{lccc}
        \toprule
        \textbf{Model} & \textbf{Prompt 1} & \textbf{Prompt 2} & \textbf{Prompt 3} \\
        \midrule
        GPT-4o mini   & 323 (64.09\%) & 415 (82.34\%) & 465 (92.26\%) \\
        GPT-4o        & 339 (67.26\%) & 414 (82.12\%) & 458 (90.87\%) \\
        DeepSeek-V3   & 263 (52.18\%) & 337 (66.86\%) & 409 (81.15\%) \\
        Sabiá-3.1       & 249 (49.4\%) & 283 (56.15\%) & 296 (58.73\%) \\
        \bottomrule
    \end{tabular}
\end{table}

\begin{table}[htbp]
    \centering
    \caption{Valid opinions incorrectly identified as hallucinations by models and prompts (false positives). Percentages are based on the 3\,732 manually annotated valid opinions.}
    \label{tab:consolidated_hallucination_results_false_positive}
    \begin{tabular}{lccc}
        \toprule
        \textbf{Model} & \textbf{Prompt 1} & \textbf{Prompt 2} & \textbf{Prompt 3} \\
        \midrule
        GPT-4o mini   & 61 (1.63\%) & 460 (12.32\%) & 1\,329 (35.59\%) \\
        GPT-4o        & 173 (4.63\%) & 458 (12.27\%) & 812 (21.75\%) \\
        DeepSeek-V3   & 68 (1.82\%) & 149 (3.99\%) & 434 (11.62\%) \\
        Sabiá-3.1       & 57 (1.54\%) & 87 (2.33\%) & 102 (2.73\%) \\
        \bottomrule
    \end{tabular}
\end{table}

The first point to observe is the effect of prompt engineering. As the prompts became more precise -- explicitly requiring evidence, emphasizing direct inference, or including strict directives -- the true positive rate increased. Consequently, for a given LLM, Prompt 1 detected the fewest hallucinations, followed by Prompt 2 and then Prompt 3.

However, as illustrated in Figure \ref{fig:fig_results_true_positive_vs_false_positve}, there is a classic trade-off between the true positive rate (correctly hallucinations detected) and the false positive rate (valid opinions labeled as hallucinations). Models and prompts that detect more hallucinations tend to make more errors when classifying valid opinions.

Regardless of the model tested, increasing the total number of detected hallucinations exponentially increases the total number of valid opinions labeled as hallucinations. For the GPT-4o, DeepSeek-V3, and Sabiá-3.1 models, this increase follows a similar exponential pattern ($\approx e^{0.06x}$, where $x$ is the percentage of true positives). However, for the GPT-4o mini model, the growth is steeper ($\approx e^{0.11x}$). In this trade-off, GPT-4o mini performed the worst, while the other three models showed comparable behavior.
 
Considering the models with similar trade-off behavior and the three prompts tested, the best performance was observed  with GPT-4o, followed by DeepSeek-V3 and Sabiá-3.1. However, since the trade-off between true positives and false positives is similar across these models, it is possible that other prompt variations would yield different overall results without changing this underlying relationship. Therefore, this information alone is not sufficient to conclude that GPT-4o would outperform the others with a different set of prompts.

Regarding the susceptibility of the models to prompt variation, the Sabiá-3.1 model was the least affected: Prompt 3 detected fewer than 10 percentage points more hallucinations than Prompt 1. In contrast, GPT-4o mini was the most susceptible, with an increase of approximately 28 percentage points in hallucination detection from Prompt 1 to Prompt 3 (at the cost of an increase of about 34 percentage points in the percentage of false positives).

The proposed method is a valid way to detect hallucinations, but its effectiveness depends on the LLM and prompt used. With Prompt 3 and the GPT-4o model (Table \ref{tab:results_gpt_4o}), 812 valid opinions were misclassified as hallucinations, while 2\,922 valid opinions remained (about 14 per transcript). Among the 504 hallucinations in the dataset, only 46 went undetected (9.13\%). In other words, the method labeled 2\,968 opinions as valid (2\,922 correct + 46 missed hallucinations), meaning that about 1.5\% of the estimated valid opinions may still be hallucinations. This value should be seen as an upper bound, since the method to detect hallucinations used only four chunks of text (i.e., some of these 46 missed hallucinations might be valid opinions supported by text outside these chunks).

The results were obtained for 12 configurations (combinations of 3 prompts and 4 models). The trade-off relationships between true and false positives found were similar in 3 of the 4 models. This suggests that the relationship is not fixed; in other words, advances in LLMs may reduce this number, improving the use of LLMs for this task.

In addition, the automatic hallucination detection in this experiment did not account for the effect of the chunks retrieved and sent to the LLMs. Performance also depends on the chunk retrieval strategy. Adjustments such as increasing the number of retrieved chunks or using more accurate embedding models could lead to better results. Nevertheless, given the probabilistic nature of the methods, it is unlikely that this trade-off can be eliminated entirely.

It is worth noting that, in this experiment, each transcript was summarized as a whole, simulating the perspective of a typical ChatGPT user. In this scenario, we identified two types of hallucinations. The first occurs when an opinion expressed by one person is attributed to the wrong speaker: for example, attributing to Bob an opinion actually voiced by Alice. The second involves factually incorrect content. In a dedicated system for this task, a better approach would be to isolate each participant’s speech and summarize it individually. This could eliminate the first type of hallucination (misattributed opinions) and likely reduce the overall number of hallucinations.

\section{Conclusion}

We introduced \thedataset, a new Portuguese dataset with 206 samples for summarizing long documents (transcripts of public hearings held by the Brazilian Chamber of Deputies). We also presented a baseline using ChatGPT to extract the main individuals involved in the hearings and their opinions from the transcripts. Furthermore, we discussed evaluation metrics for this task, with a focus on detecting hallucinations in the generated summaries. This discussion resulted in 4\,238 manually annotated samples, which can be used to evaluate natural language inference systems in Portuguese.

There is a lack of summarization datasets in Portuguese, and the existing ones for long-document summarization do not include summaries of the main discussions presented in source texts. \thedataset addresses this gap and, in addition, provides summaries that have been manually verified.

We experimented with the dataset using ChatGPT and a custom prompt to summarize public hearings. The goal is not to solve the task, but to establish a baseline using a tool readily available to end users, supporting future research.

In the experiment, which focused on extracting individuals' opinions, ChatGPT achieved an average recall of nearly 45\%, meaning it retrieved almost half of the opinions present in the reference dataset. The average precision was approximately 25\% per sample: for each relevant opinion, the system generated three others not present in the reference. We manually verified the 4\,238 generated opinions and checked whether they could be inferred from the four closest chunks of text. This analysis indicated that nearly 12\% of the generated opinions may be hallucinations.

We proposed a method for hallucination detection that uses an LLM as a judge. Our findings reveal a clear trade-off: the more hallucinations are detected, the more valid opinions are labeled as hallucinations. We analyzed this relationship across three prompts and found that, for the GPT-4o, DeepSeek-V3, and Sabiá-3.1 models, the increase in valid opinions labeled as hallucinations follows a similar exponential pattern ($\approx e^{0.06x}$, where $x$ is the percentage of hallucinations detected). In contrast, for the GPT-4o mini model, the growth is steeper: $\approx e^{0.11x}$.

The experiment yielded relevant insights and suggests that a real-world summarization system needs to be more complex than current off-the-shelf tools allow. As a direction for future research, we propose designing a dedicated summarization system for public hearings where the process is divided into stages, starting with the separation of each participant's speech to reduce the risk of hallucinations. Other possibilities include using the dataset to automatically generate news articles based on either the metadata or the full transcript of the hearing.

\bibliography{main.bib}

\FloatBarrier

\appendix

\section{Original prompts in Portuguese}
\label{sec:prompts_pt}

This appendix presents all the prompts used in this article, in Portuguese. Appendix \ref{sec:prompts_en} contains their English translations.

\begin{figure}[!htb]
\begin{tcolorbox}[style_prompt]
\textbf{[System]}

Você é um assistente que analisa matérias escritas pela Agência Câmara, da Câmara dos Deputados. Seu papel é identificar na matéria os seguintes itens:\\

- Tópico principal que está sendo tratado\\
- O nome das pessoas envolvidas\\
- O que cada pessoa defende\\
- O que cada pessoa disse (em caso de existir citação direta)\\

Desconsidere o nome dos jornalistas ou editores da matéria. As únicas pessoas que interessam são as que estão no corpo da matéria.\\

O retorno deve ser no formato JSON, com duas propriedades: \\

-- "assunto": uma string que indica o assunto principal da matéria\\
-- "envolvidos": uma lista de objetos que indica as pessoas envolvidas na matéria. O objeto deve ter três propriedades:\\
\hspace*{2em} -- "nome": string, indica nome da pessoa\\
\hspace*{2em} -- "cargo": string, indica cargo que a pessoa ocupa, juntamente com o órgão, a entidade ou a empresa em que ela trabalha, se estiver disponível\\
\hspace*{2em} -- "opinioes": lista de string indicando todas as opiniões que a pessoa defendeu e que estão indicadas no texto. As opiniões devem ser listadas de forma detalhada. Se for uma citação direta, o texto indicado na lista DEVE OBRIGATORIAMENTE ser idêntico ao contido na matéria, incluindo as aspas.\\

\textbf{[User]}

\{ARTICLE\_TEXT\}

\end{tcolorbox}
\caption{Prompt used to extract the metadata of the news articles used in \thedataset.}
\label{fig:prompt_extract_metadata_from_articles}
\end{figure}

\begin{figure}[!htb]
\begin{tcolorbox}[style_prompt]
O usuário irá enviar uma transcrição de uma audiência pública realizada em alguma Comissão da Câmara dos Deputados. Seu papel é ler o arquivo que será enviado e identificar os seguintes itens:\\

-- Tópico principal da audiência pública\\
-- O nome das pessoas envolvidas\\
-- O que cada pessoa defende ou comenta\\

Os itens identificados devem possibilitar que o usuário redija uma matéria jornalística com início, meio e fim.\\

A transcrição contém a fala exata dita pelos participantes. Inicialmente o participante é identificado. Todo o texto que se segue até a identificação de uma nova pessoa é a fala daquele participante. O texto possui o seguinte formato:\\

\texttt{[[O(A) SR.(SRA.) PESSOA 1]]}\\
\texttt{[[UM OU MAIS PARÁGRAFOS CONTENDO TODA A FALA DA PESSOA 1]]}\\

\texttt{[[O(A) SR.(SRA.) PESSOA 2]]}\\
\texttt{[[UM OU MAIS PARÁGRAFOS CONTENDO TODA A FALA DA PESSOA 2]]}\\

...\\
...\\

\texttt{[[O(A) SR.(SRA.) PESSOA N]]}\\
\texttt{[[PARÁGRAFOS CONTENDO TODA A FALA DA PESSOA N]]}\\

Após ler e analisar o documento enviado, você deverá dar a sua resposta no formato JSON com três propriedades: \\

-- "assunto": uma string que indica o assunto principal da audiência pública. Essa informação normalmente está na primeira fala do primeiro participante, logo na abertura da audiência.\\
-- "envolvidos": uma lista de objetos que indica as pessoas envolvidas no debate. O objeto deve ter três propriedades:\\
\hspace*{2em} -- "nome": string, indica nome da pessoa\\
\hspace*{2em} -- "cargo": string, indica cargo que a pessoa ocupa, juntamente com o órgão, a entidade ou a empresa em que ela trabalha, se estiver disponível\\
\hspace*{2em} -- "opinioes": lista de string indicando TODAS as opiniões relevantes ao assunto que a pessoa defendeu e que estão indicadas no texto. As opiniões devem ser listadas de forma detalhada\\
-- "tl\_dr": um resumo que possibilitará ao usuário escrever uma matéria jornalista sobre a audiência pública usando os dados (nome e opiniões) dos envolvidos extraídos da transcrição\\

\end{tcolorbox}
\caption{Instruct prompt of the custom GPT used to extract the metadata of the transcripts in \thedataset.}
\label{fig:prompt_instruct_custom_gpt}
\end{figure}

\begin{figure}[!htb]
\begin{tcolorbox}[style_prompt]
\textbf{[System]}

Você receberá duas listas de pessoas, com seus nomes e cargos.\\
Uma mesma pessoa pode aparecer nas duas listas, mas de maneira diferente (pequenas alterações no nome, variação do cargo...)\\
Seu objetivo é criar uma correspondência entre as duas listas, no formato de um dicionário, em que as chaves são o nome da pessoa na lista 1, e os valores são os nomes correspondentes da lista 2. Anote apenas os nomes, e não os cargos. E se não houver uma correspondência, escreva 'Não identificado'.\\
Sua resposta final deve ser apenas o JSON, não escreva nada além disso.\\

Exemplo\\
Lista 1:\\
Nome: João Carlos Souza. Cargo: Engenheiro\\
Nome: Júlia da Silva Macedo. Cargo: Jornalista investigativa\\
Nome: Lucas Ferreira. Cargo: Assessor de imprensa\\

Lista 2:\\
Nome: Pablo Machado. Cargo: Psicólogo\\
Nome: João Carlos de Souza. Cargo: Engenheiro de Produção\\
Nome: Júlia Macedo. Cargo: Jornalista\\

Resposta:\\
\{\\
\hspace*{2em} "João Carlos Souza": "João Carlos de Souza",\\
\hspace*{2em} "Júlia da Silva Macedo": "Júlia Macedo",\\
\hspace*{2em} "Lucas Ferreira": "Não identificado"\\
\}\\

\end{tcolorbox}
\caption{Prompt to map individuals in two lists (fictitious names).}
\label{fig:prompt_match_individuals}
\end{figure}

\begin{figure}[!htb]
\begin{tcolorbox}[style_prompt]
\textbf{[System]}

Você receberá duas listas contendo assuntos tratados em uma audiência pública na Câmara dos Deputados.\\
Um mesmo assunto pode aparecer nas duas listas, mas escrito de formas diferentes. Seu objetivo é avaliar o conteúdo dos assuntos e fazer um mapeamento entre cada item da primeira lista com itens da segunda, indicando quais são similares.\\
Para serem considerados similares, os assuntos devem necessariamente representar uma mesma informação, mas escrita de diferentes formas.\\
Os assuntos de cada lista serão numerados, e você deve criar um dicionário com o mapeamento dos assuntos de cada lista.\\
Um mesmo assunto da segunda lista pode ser mapeado para mais de um assunto da lista 1, e vice-versa.\\
Os assuntos sem um correspondente devem ser marcados com "Não identificado".\\
Sua resposta deve ser apenas o JSON, não escreva nada além disso.\\

Exemplo:\\
Lista 1:\\
1. Acusou o deputado Fulano de publicar mentiras em redes sociais, e defendeu que o mesmo seja punido por isso.\\
2. Destacou a importância de haver verificações de veracidade de publicações em todas as redes sociais.\\
3. Defendeu o bloqueio de contas que publiquem informações falsas.\\

Lista 2:\\
1. Citou o Projeto de Lei 2630/20, conhecido como PL das Fake News.\\
2. Fez uma acusação contra o deputado Fulano, por publicar fake news em seu facebook.\\
3. Insinuou que o deputado Fulano seja punido.\\
4. Insinuou que contas que publiquem fake news devem ser derrubadas das redes sociais.\\
5. Citou o nome de envolvidos no Projeto de Lei.\\

Resposta:\\
\{\\
\hspace*{2em} "1": "2, 3",\\
\hspace*{2em} "2": "Não identificado",\\
\hspace*{2em} "3": "4"\\
\}

\end{tcolorbox}
\caption{Prompt to map opinions.}
\label{fig:prompt_match_opinions}
\end{figure}

\begin{figure}[!htb]
\begin{tcolorbox}[style_prompt]
\textbf{[System]}\\
Você é um assistente que analisa se uma opinião pode ser completamente inferida a partir de um texto.\\

O retorno da sua análise deverá ser sempre no formato JSON e conterá duas propriedades:\\
\hspace*{2em}- "explicacao": Uma string com o seu raciocínio explicando o porque a opinião pode ou não ser inferida pelo texto;\\
\hspace*{2em}- "opiniao\_inferida": Um boolean (true ou false) sintetizando sua explicação: true, se a opinião puder ser inferida a partir do texto, ou false, se não puder.\\

Não forneça nada além do JSON com as propriedades acima.\\

\textbf{[User]}\\
\#\#\#\#\#\# TEXTO:\\
\{TEXTO\}\\

\#\#\#\#\#\# OPINIÃO PARA ANALISAR:\\
\{OPINIAO\}

\end{tcolorbox}
\caption{Prompt 1 to detect hallucinations.}
\label{fig:prompt_1_hallucination}
\end{figure}

\begin{figure}[!htb]
\begin{tcolorbox}[style_prompt]
\textbf{[System]}\\
Você é um assistente especializado em análise de texto. Sua tarefa é verificar se uma opinião pode ser COMPLETAMENTE inferida a partir de um texto fornecido. Para isso, siga as etapas abaixo:\\

1. Identifique, no texto, os trechos que podem servir para suportar a opinião analisada;\\
2. Verifique se TODA a opinião é suportada pelos trechos selecionados. NÃO FAÇA SUPOSIÇÕES;\\
3. Forneça uma resposta direta (boolean). A resposta deve indicar se TODA a opinião pode ser inferida DIRETAMENTE do texto, sem o uso de suposições.\\

O retorno da sua análise deverá ser sempre no formato JSON e conterá três propriedades referentes aos passos anteriores:\\
\hspace*{2em}- "trechos\_para\_basear\_analise": Uma lista de strings com os trechos que podem servir para suportar a opinião;\\
\hspace*{2em}- "explicacao": Uma string com o seu raciocínio explicando o porque a opinião pode ou não ser inferida pelo texto;\\
\hspace*{2em}- "opiniao\_inferida": Um boolean (true ou false) sintetizando sua explicação: true, se a opinião puder ser inferida a partir do texto, ou false, se não puder.\\
    
Não forneça nada além do JSON com as propriedades acima.\\

\textbf{[User]}\\
\#\#\#\#\#\# TEXTO:\\
\{TEXTO\}\\

\#\#\#\#\#\# OPINIÃO PARA ANALISAR:\\
\{OPINIAO\}

\end{tcolorbox}
\caption{Prompt 2 to detect hallucinations.}
\label{fig:prompt_2_hallucination}
\end{figure}

\begin{figure}[!htb]
\begin{tcolorbox}[style_prompt]
\textbf{[System]}\\
Você é um assistente especializado em análise de discursos. Sua tarefa é verificar se uma opinião pode ser COMPLETAMENTE inferida a partir de um trechos de texto. A sua análise deve seguir as etapas abaixo:\\

1. Identifique, nos trechos de texto, frases que suportam a opinião analisada;\\
2. Verifique se TODA a opinião é suportada pelas frases selecionadas. Não faça suposições e inferências indiretas;\\
3. Forneça uma resposta direta (boolean). A resposta deve indicar se TODA a opinião pode ser inferida DIRETAMENTE do texto. Caso você tenha dúvidas ou apenas parte da opinião puder ser inferida, responda que a opinião não pode ser inferida.\\

O retorno da sua análise deverá ser sempre no formato JSON e conterá três propriedades referentes aos passos anteriores:\\
\hspace*{2em}- "trechos\_para\_basear\_analise": Uma lista com potenciais frases que suportam a opinião;\\
\hspace*{2em}- "explicacao": Uma string com o seu raciocínio explicando se TODA a opinião pode ou não ser inferida pelo texto;\\
\hspace*{2em}- "opiniao\_inferida": Um boolean (true ou false) sintetizando sua explicação: true, se TODA a opinião puder ser inferida a partir do texto, ou false, se não puder ou se for inconclusivo.\\
    
Não forneça nada além do JSON com as propriedades acima.\\

\textbf{[User]}\\
\#\#\#\#\#\# TEXTO:\\
\{TEXTO\}\\

\#\#\#\#\#\# OPINIÃO PARA ANALISAR:\\
\{OPINIAO\}

\end{tcolorbox}
\caption{Prompt 3 to detect hallucinations.}
\label{fig:prompt_3_hallucination}
\end{figure}

\FloatBarrier

\section{Translated prompts into English}
\label{sec:prompts_en}

This appendix presents the English translations of the prompts shown in Appendix \ref{sec:prompts_pt}.

\begin{figure}[!htb]
\begin{tcolorbox}[style_prompt]
\textbf{[System]}

You are an assistant who analyzes news articles written by the Agência Câmara, from the Chamber of Deputies. Your role is to identify the following items in the article:\\

- Main topic being addressed\\
- Names of the people involved\\
- What each person advocates\\
- What each person said (in case of direct quotation)\\

Ignore the names of the journalists or editors of the article. Only those mentioned in the body of the article should be considered.\\

The response should be in JSON format with two properties: \\

-- "assunto": a string that indicates the main subject of the article\\
-- "envolvidos": a list of objects that indicates the people involved in the article. Each object should have three properties:\\
\hspace*{2em} -- "nome": string, indicating the person's name\\
\hspace*{2em} -- "cargo": string, indicating the position held by the person, along with the agency, organization, or company where they work for, if available\\
\hspace*{2em} -- "opinioes": a list of strings indicating all the opinions expressed by the person that are mentioned in the text. The opinions must be described in detail. In the case of a direct quotation, the text in the list MUST match the article exactly, including quotation marks.\\

\textbf{[User]}

\{ARTICLE\_TEXT\}

\end{tcolorbox}
\caption{Prompt used to extract the metadata of the news articles used in \thedataset (translation of Figure \ref{fig:prompt_extract_metadata_from_articles}).}
\label{fig:prompt_extract_metadata_from_articles_en}
\end{figure}

\begin{figure}[!htb]
\begin{tcolorbox}[style_prompt]
The user will send a transcript of a public hearing held in a Committee of the Chamber of Deputies. Your role is to read the file that will be sent and identify the following items:\\

-- Main topic of the public hearing\\
-- The names of the people involved\\
-- What each person advocates or comments on\\

The identified items should make it possible for the user to write a well-structured journalistic article.\\

The transcript contains the exact speech given by the participants. Initially, the participant is identified. All the text that follows until the identification of a new person is the speech of that participant. The text has the following format:\\

\texttt{[[O(A) SR.(SRA.) PERSON 1]]}\\
\texttt{[[ONE OR MORE PARAGRAPHS CONTAINING THE ENTIRE SPEECH OF PERSON 1]]}\\

\texttt{[[O(A) SR.(SRA.) PERSON 2]]}\\
\texttt{[[ONE OR MORE PARAGRAPHS CONTAINING THE ENTIRE SPEECH OF PERSON 1]]}\\

...\\
...\\

\texttt{[[O(A) SR.(SRA.) PERSON N]]}\\
\texttt{[[ONE OR MORE PARAGRAPHS CONTAINING THE ENTIRE SPEECH OF PERSON N]]}\\

After reading and analyzing the submitted document, you should provide your response in JSON format with three properties: \\

-- "assunto": a string that indicates the main subject of the public hearing. This information is usually found in the first speech of the first participant, right at the opening of the hearing.\\
-- "envolvidos": a list of objects that indicates the people involved in the debate. Each object should have three properties:\\
\hspace*{2em} -- "nome": string, indicating the person's name\\
\hspace*{2em} -- "cargo": string, indicates the position the person holds, together with the agency, organization, or company where they work, if available.\\
\hspace*{2em} -- "opinioes": a list of strings indicating ALL the relevant opinions about the subject discussed in the hearing that are are mentioned in the text. The opinions must be listed in detail\\
-- "tl\_dr": a summary that allows the user to draft a news article about the public hearing based on the participants' names and opinions extracted from the transcript\\

\end{tcolorbox}
\caption{Instruct prompt of the custom GPT used to extract the metadata of the transcripts in \thedataset (translation of Figure \ref{fig:prompt_instruct_custom_gpt}).}
\label{fig:prompt_instruct_custom_gpt_en}
\end{figure}

\begin{figure}[!htb]
\begin{tcolorbox}[style_prompt]
\textbf{[System]}

You will receive two lists of people, with their names and positions.\\
The same person may appear in both lists, but with slight differences (such as small variations in the name or position).\\
Your goal is to create a mapping between the two lists in the form of a dictionary, where the keys are names from list 1, and the values are the corresponding names from list 2. Write only the names, ignore the positions. If there is no match, write 'Not identified'.\\
Your final response should be only the JSON; do not write anything else.\\

Example\\
List 1:\\
Name: João Carlos Souza. Position: Engineer\\
Name: Júlia da Silva Macedo. Position: Investigative journalist\\
Name: Lucas Ferreira. Position: Press officer\\

List 2:\\
Name: Pablo Machado. Position: Psychologist\\
Name: João Carlos de Souza. Position: Production Engineer\\
Name: Júlia Macedo. Position: Journalist\\

Answer:\\
\{\\
\hspace*{2em} "João Carlos Souza": "João Carlos de Souza",\\
\hspace*{2em} "Júlia da Silva Macedo": "Júlia Macedo",\\
\hspace*{2em} "Lucas Ferreira": "Not identified"\\
\}\\

\end{tcolorbox}
\caption{Prompt to map individuals in two lists (fictitious names, translation of Figure \ref{fig:prompt_match_individuals}).}
\label{fig:prompt_match_individuals_en}
\end{figure}

\begin{figure}[!htb]
\begin{tcolorbox}[style_prompt]
\textbf{[System]}

You will receive two lists containing topics discussed in a public hearing in the Chamber of Deputies.\\
The same topic may appear in both lists, but written in different ways. Your goal is to evaluate the content of the topics and create a mapping between each item in the first list and items in the second, indicating which ones are similar.\\
To be considered similar, the topics must necessarily represent the same information but written in different ways.\\
The topics in each list will be numbered, and you should create a dictionary mapping the topics from each list.\\
The same topic from the second list can be mapped to more than one topic from the first list, and vice versa.\\
Topics without a corresponding match should be marked with "Not identified."\\
Your final response should be only the JSON; do not write anything else.\\

Example:\\
List 1:\\
1. Accused Deputy John Doe of publishing lies on social media and advocated for him to be punished for it.\\
2. Highlighted the importance of having fact-checking for publications on all social media platforms.\\
3. Advocated for the blocking of accounts that publish false information.\\

List 2:\\
1. Mentioned Bill 2630/20, known as the Fake News Bill.\\
2. Made an accusation against Deputy John Doe for publishing fake news on his Facebook.\\
3. Insinuated that Deputy John Doe should be punished.\\
4. Insinuated that accounts that publish fake news should be taken down from social media.\\
5. Mentioned the names of those involved in the Bill.\\

Answer:\\
\{\\
\hspace*{2em} "1": "2, 3",\\
\hspace*{2em} "2": "Not identified",\\
\hspace*{2em} "3": "4"\\
\}

\end{tcolorbox}
\caption{Prompt to map opinions (translation of Figure \ref{fig:prompt_match_opinions}).}
\label{fig:prompt_match_opinions_en}
\end{figure}

\begin{figure}[!htb]
\begin{tcolorbox}[style_prompt]
\textbf{[System]}\\
You are an assistant who analyzes whether an opinion can be completely inferred from a text.\\

The output of your analysis should always be in JSON format and will contain two properties:\\
\hspace*{2em}- "explicacao": A string with your reasoning explaining why the opinion can or cannot be inferred from the text;\\
\hspace*{2em}- "opiniao\_inferida": A boolean (true or false) summarizing your explanation: true, if the opinion can be inferred from the text, or false, if it cannot.\\

Do not provide anything other than the JSON with the properties above.\\

\textbf{[User]}\\
\#\#\#\#\#\# TEXT:\\
\{TEXT\}\\

\#\#\#\#\#\# OPINION TO BE ANALYZE:\\
\{OPINION\}

\end{tcolorbox}
\caption{Prompt 1 to detect hallucinations (translation of Figure \ref{fig:prompt_1_hallucination}).}
\label{fig:prompt_1_hallucination_en}
\end{figure}

\begin{figure}[!htb]
\begin{tcolorbox}[style_prompt]
\textbf{[System]}\\
You are an assistant specialized in text analysis. Your task is to verify whether an opinion can be COMPLETELY inferred from a provided text. To do this, follow the steps below:\\

1. Identify, in the text, the excerpts that can support the analyzed opinion;\\
2. Verify whether the ENTIRE opinion is supported by the selected excerpts. DO NOT MAKE ASSUMPTIONS;\\
3. Provide a direct (boolean) response. The answer should indicate whether the ENTIRE opinion can be INFERRED DIRECTLY from the text, without making assumptions.\\

The output of your analysis should always be in JSON format and will contain three properties related to the previous steps:\\
\hspace*{2em}- "trechos\_para\_basear\_analise": A list of strings with the excerpts that can support the opinion;\\
\hspace*{2em}- "explicacao": A string with your reasoning explaining why the opinion can or cannot be inferred from the text;\\
\hspace*{2em}- "opiniao\_inferida": A boolean (true or false) summarizing your explanation: true, if the opinion can be inferred from the text, or false, if it cannot.\\
    
Do not provide anything other than the JSON with the properties above.\\

\textbf{[User]}\\
\#\#\#\#\#\# TEXT:\\
\{TEXT\}\\

\#\#\#\#\#\# OPINION TO BE ANALYZE:\\
\{OPINION\}

\end{tcolorbox}
\caption{Prompt 2 to detect hallucinations (translation of \ref{fig:prompt_2_hallucination}).}
\label{fig:prompt_2_hallucination_en}
\end{figure}

\begin{figure}[!htb]
\begin{tcolorbox}[style_prompt]
\textbf{[System]}\\
You are an assistant specialized in discourse analysis. Your task is to verify whether an opinion can be COMPLETELY inferred from a text excerpt. Your analysis should follow the steps below:\\

1. Identify, in the text excerpts, sentences that support the analyzed opinion;\\
2. Verify if the ENTIRE opinion is supported by the selected sentences. Do not make assumptions or indirect inferences;\\
3. Provide a direct (boolean) answer. The answer should indicate whether the ENTIRE opinion can be INFERRED DIRECTLY from the text. If you have doubts or only part of the opinion can be inferred, respond that the opinion cannot be inferred.\\

The output of your analysis should always be in JSON format and will contain three properties related to the previous steps:\\
\hspace*{2em}- "trechos\_para\_basear\_analise": A list of potential sentences that support the opinion;\\
\hspace*{2em}- "explicacao": A string with your reasoning explaining whether the ENTIRE opinion can or cannot be inferred from the text;\\
\hspace*{2em}- "opiniao\_inferida": A boolean (true or false) summarizing your explanation: true, if the ENTIRE opinion can be inferred from the text, or false, if it cannot or if it is inconclusive.\\
    
Do not provide anything other than the JSON with the properties above.\\

\textbf{[User]}\\
\#\#\#\#\#\# TEXT:\\
\{TEXT\}\\

\#\#\#\#\#\# OPINION TO BE ANALYZE:\\
\{OPINION\}

\end{tcolorbox}
\caption{Prompt 3 to detect hallucinations (translation of Figure \ref{fig:prompt_3_hallucination}).}
\label{fig:prompt_3_hallucination_en}
\end{figure}

\FloatBarrier

\section{Confusion matrices for the experiment}
\label{sec:confusion_matrices}

\begin{table}[htbp]
\caption{Confusion matrices for automatic hallucination detection using GPT-4o mini.}
\centering
{%
\begin{tabular}{l|l|c|c|c}

\multicolumn{2}{c}{}&\multicolumn{2}{c}{Prompt 1}&\\
\cmidrule{3-4}
\multicolumn{2}{c|}{}&Valid opinions&Hallucination&\multicolumn{1}{c}{Total}\\
\cmidrule{2-4}
\multirow{2}{*}{Dataset}&Valid opinions& 3\,673 & 61 & 3\,734\\
\cmidrule{2-4}
& Hallucination & 181 & 323 & 504 \\
\cmidrule{2-4}
\multicolumn{1}{c}{} & \multicolumn{1}{c}{Total} & \multicolumn{1}{c}{3\,854} & \multicolumn{1}{c}{384} & \multicolumn{1}{c}{4\,238}\\
\end{tabular}
}
\vspace{0.05cm}

\centering
{%
\begin{tabular}{l|l|c|c|c}

\multicolumn{2}{c}{}&\multicolumn{2}{c}{Prompt 2}&\\
\cmidrule{3-4}
\multicolumn{2}{c|}{}&Valid opinions&Hallucination&\multicolumn{1}{c}{Total}\\
\cmidrule{2-4}
\multirow{2}{*}{Dataset}&Valid opinions& 3\,274 & 460 & 3\,734\\
\cmidrule{2-4}
& Hallucination & 89 & 415 & 504 \\
\cmidrule{2-4}
\multicolumn{1}{c}{} & \multicolumn{1}{c}{Total} & \multicolumn{1}{c}{3\,363} & \multicolumn{1}{c}{875} & \multicolumn{1}{c}{4\,238}\\
\end{tabular}
}
\vspace{0.05cm}

{%
\begin{tabular}{l|l|c|c|c}

\multicolumn{2}{c}{}&\multicolumn{2}{c}{Prompt 3}&\\
\cmidrule{3-4}
\multicolumn{2}{c|}{}&Valid opinions&Hallucination&\multicolumn{1}{c}{Total}\\
\cmidrule{2-4}
\multirow{2}{*}{Dataset}&Valid opinions& 2\,405 & 1\,329 & 3\,734\\
\cmidrule{2-4}
& Hallucination & 39 & 465 & 504 \\
\cmidrule{2-4}
\multicolumn{1}{c}{} & \multicolumn{1}{c}{Total} & \multicolumn{1}{c}{2\,444} & \multicolumn{1}{c}{1\,974} & \multicolumn{1}{c}{4\,238}\\
\end{tabular}
}
\label{tab:results_gpt_4o_mini}
\end{table}

\begin{table}[htbp]
\caption{Confusion matrices for automatic hallucination detection using GPT-4o.}
\centering
{%
\begin{tabular}{l|l|c|c|c}

\multicolumn{2}{c}{}&\multicolumn{2}{c}{Prompt 1}&\\
\cmidrule{3-4}
\multicolumn{2}{c|}{}&Valid opinions&Hallucination&\multicolumn{1}{c}{Total}\\
\cmidrule{2-4}
\multirow{2}{*}{Dataset}&Valid opinions& 3\,561 & 173 & 3\,734\\
\cmidrule{2-4}
& Hallucination & 165 & 339 & 504 \\
\cmidrule{2-4}
\multicolumn{1}{c}{} & \multicolumn{1}{c}{Total} & \multicolumn{1}{c}{3\,726} & \multicolumn{1}{c}{512} & \multicolumn{1}{c}{4\,238}\\
\end{tabular}
}
\vspace{0.05cm}

{%
\begin{tabular}{l|l|c|c|c}

\multicolumn{2}{c}{}&\multicolumn{2}{c}{Prompt 2}&\\
\cmidrule{3-4}
\multicolumn{2}{c|}{}&Valid opinions&Hallucination&\multicolumn{1}{c}{Total}\\
\cmidrule{2-4}
\multirow{2}{*}{Dataset}&Valid opinions& 3\,276 & 458 & 3\,734\\
\cmidrule{2-4}
& Hallucination & 90 & 414 & 504 \\
\cmidrule{2-4}
\multicolumn{1}{c}{} & \multicolumn{1}{c}{Total} & \multicolumn{1}{c}{3\,366} & \multicolumn{1}{c}{872} & \multicolumn{1}{c}{4\,238}\\
\end{tabular}
}
\vspace{0.05cm}

{%
\begin{tabular}{l|l|c|c|c}

\multicolumn{2}{c}{}&\multicolumn{2}{c}{Prompt 3}&\\
\cmidrule{3-4}
\multicolumn{2}{c|}{}&Valid opinions&Hallucination&\multicolumn{1}{c}{Total}\\
\cmidrule{2-4}
\multirow{2}{*}{Dataset}&Valid opinions& 2\,922 & 812 & 3\,734\\
\cmidrule{2-4}
& Hallucination & 46 & 458 & 504 \\
\cmidrule{2-4}
\multicolumn{1}{c}{} & \multicolumn{1}{c}{Total} & \multicolumn{1}{c}{2\,968} & \multicolumn{1}{c}{1\,270} & \multicolumn{1}{c}{4\,238}\\
\end{tabular}
}

\label{tab:results_gpt_4o}
\end{table}

\begin{table}[htbp]
\caption{Confusion matrix for automatic hallucination detection using DeepSeek-V3.}
\centering
{%
\begin{tabular}{l|l|c|c|c}

\multicolumn{2}{c}{}&\multicolumn{2}{c}{Prompt 1}&\\
\cmidrule{3-4}
\multicolumn{2}{c|}{}&Valid opinions&Hallucination&\multicolumn{1}{c}{Total}\\
\cmidrule{2-4}
\multirow{2}{*}{Dataset}&Valid opinions& 3\,666 & 68 & 3\,734\\
\cmidrule{2-4}
& Hallucination & 241 & 263 & 504 \\
\cmidrule{2-4}
\multicolumn{1}{c}{} & \multicolumn{1}{c}{Total} & \multicolumn{1}{c}{2\,968} & \multicolumn{1}{c}{1\,270} & \multicolumn{1}{c}{4\,238}\\
\end{tabular}
}
\vspace{0.05cm}

{%
\begin{tabular}{l|l|c|c|c}

\multicolumn{2}{c}{}&\multicolumn{2}{c}{Prompt 2}&\\
\cmidrule{3-4}
\multicolumn{2}{c|}{}&Valid opinions&Hallucination&\multicolumn{1}{c}{Total}\\
\cmidrule{2-4}
\multirow{2}{*}{Dataset}&Valid opinions& 3\,585 & 149 & 3\,734\\
\cmidrule{2-4}
& Hallucination & 167 & 337 & 504 \\
\cmidrule{2-4}
\multicolumn{1}{c}{} & \multicolumn{1}{c}{Total} & \multicolumn{1}{c}{2\,968} & \multicolumn{1}{c}{1\,270} & \multicolumn{1}{c}{4\,238}\\
\end{tabular}
}
\vspace{0.05cm}

{%
\begin{tabular}{l|l|c|c|c}

\multicolumn{2}{c}{}&\multicolumn{2}{c}{Prompt 3}&\\
\cmidrule{3-4}
\multicolumn{2}{c|}{}&Valid opinions&Hallucination&\multicolumn{1}{c}{Total}\\
\cmidrule{2-4}
\multirow{2}{*}{Dataset}&Valid opinions& 3\,300 & 434 & 3\,734\\
\cmidrule{2-4}
& Hallucination & 95 & 409 & 504 \\
\cmidrule{2-4}
\multicolumn{1}{c}{} & \multicolumn{1}{c}{Total} & \multicolumn{1}{c}{2\,968} & \multicolumn{1}{c}{1\,270} & \multicolumn{1}{c}{4\,238}\\
\end{tabular}
}

\label{tab:results_deepseek}
\end{table}

\begin{table}[htbp]
\caption{Confusion matrix for automatic hallucination detection using Sabiá-3.1.}
\centering
{%
\begin{tabular}{l|l|c|c|c}

\multicolumn{2}{c}{}&\multicolumn{2}{c}{Prompt 1}&\\
\cmidrule{3-4}
\multicolumn{2}{c|}{}&Valid opinions&Hallucination&\multicolumn{1}{c}{Total}\\
\cmidrule{2-4}
\multirow{2}{*}{Dataset}&Valid opinions& 3\,677 & 57 & 3\,734\\
\cmidrule{2-4}
& Hallucination & 255 & 249 & 504 \\
\cmidrule{2-4}
\multicolumn{1}{c}{} & \multicolumn{1}{c}{Total} & \multicolumn{1}{c}{2\,968} & \multicolumn{1}{c}{1\,270} & \multicolumn{1}{c}{4\,238}\\
\end{tabular}
}
\vspace{0.05cm}

{%
\begin{tabular}{l|l|c|c|c}

\multicolumn{2}{c}{}&\multicolumn{2}{c}{Prompt 2}&\\
\cmidrule{3-4}
\multicolumn{2}{c|}{}&Valid opinions&Hallucination&\multicolumn{1}{c}{Total}\\
\cmidrule{2-4}
\multirow{2}{*}{Dataset}&Valid opinions& 3\,647 & 87 & 3\,734\\
\cmidrule{2-4}
& Hallucination & 221 & 283 & 504 \\
\cmidrule{2-4}
\multicolumn{1}{c}{} & \multicolumn{1}{c}{Total} & \multicolumn{1}{c}{2\,968} & \multicolumn{1}{c}{1\,270} & \multicolumn{1}{c}{4\,238}\\
\end{tabular}
}
\vspace{0.05cm}

{%
\begin{tabular}{l|l|c|c|c}

\multicolumn{2}{c}{}&\multicolumn{2}{c}{Prompt 3}&\\
\cmidrule{3-4}
\multicolumn{2}{c|}{}&Valid opinions&Hallucination&\multicolumn{1}{c}{Total}\\
\cmidrule{2-4}
\multirow{2}{*}{Dataset}&Valid opinions& 3\,632 & 102 & 3\,734\\
\cmidrule{2-4}
& Hallucination & 208 & 296 & 504 \\
\cmidrule{2-4}
\multicolumn{1}{c}{} & \multicolumn{1}{c}{Total} & \multicolumn{1}{c}{2\,968} & \multicolumn{1}{c}{1\,270} & \multicolumn{1}{c}{4\,238}\\
\end{tabular}
}

\label{tab:results_sabia}
\end{table}

\end{document}